\def\year{2021}\relax
\documentclass[letterpaper]{article}

\usepackage{aaai21}       % DO NOT CHANGE THIS
\usepackage{times}        % DO NOT CHANGE THIS
\usepackage{helvet}       % DO NOT CHANGE THIS
\usepackage{courier}      % DO NOT CHANGE THIS
\usepackage[hyphens]{url} % DO NOT CHANGE THIS
\usepackage{graphicx}     % DO NOT CHANGE THIS
\usepackage{natbib}       % DO NOT CHANGE THIS AND DO NOT ADD ANY OPTIONS TO IT
\usepackage{caption}      % DO NOT CHANGE THIS AND DO NOT ADD ANY OPTIONS TO IT
\usepackage{amsmath}
\usepackage{amssymb}
\usepackage{algorithmic}
\usepackage{multirow}
\usepackage{makecell}
\usepackage{colortbl}
\usepackage{bbding}
\usepackage{pifont}
\usepackage[caption=false]{subfig}
\usepackage{algorithm}
\usepackage{bm}

\graphicspath{{./Imgs/}}
\DeclareGraphicsExtensions{.pdf,.jpg,.png}

\def\vs{\textit{vs.~}}
\def\ie{\textit{i.e.,~}}
\def\eg{\textit{e.g.,~}}

\def\sArt{{state-of-the-art~}}

\newcommand{\figref}[1]{Fig.~\ref{#1}}
\newcommand{\tabref}[1]{Tab.~\ref{#1}}
\newcommand{\equref}[1]{Eqn.~\ref{#1}}

\newcommand{\myPara}[1]{\vspace{.12in}\noindent\textbf{#1}}
\definecolor{mycolor}{rgb}{.9,.9,.9}

\newcommand{\tabincell}[2]{\begin{tabular}{@{}#1@{}}#2\end{tabular}}

\title{MiniSeg: An Extremely Minimum Network for Efficient COVID-19 Segmentation}

\iftrue%\iffalse
\author{
    Yu Qiu\textsuperscript{\rm 1},
    Yun Liu\textsuperscript{\rm 2*},
    Shijie Li\textsuperscript{\rm 3},
    Jing Xu\textsuperscript{\rm 1}\thanks{Yun Liu and Jing Xu are corresponding authors.} \\
}
\affiliations{
    \textsuperscript{\rm 1} College of Artificial Intelligence, Nankai University \\
    \textsuperscript{\rm 2} College of Computer Science, Nankai University \\
    \textsuperscript{\rm 3} Department of Information Systems and Artificial Intelligence, University of Bonn \\
    yqiu@mail.nankai.edu.cn, 
    \{vagrantlyun, jason.li.nku\}@gmail.com, 
    xujing@nankai.edu.cn
}
\else
\author{Anonymous AAAI submission \\
Paper ID 86 \\
}
\fi

\begin{document}

\maketitle

\begin{abstract}
The rapid spread of the new pandemic, \ie COVID-19, 
has severely threatened global health.
Deep-learning-based computer-aided screening, \eg COVID-19 infected 
CT area segmentation, has attracted much attention.
However, the publicly available COVID-19 training data are limited, 
easily causing overfitting for traditional deep learning 
methods that are usually data-hungry with millions of parameters.
On the other hand, fast training/testing and low computational 
cost are also necessary for quick deployment and development 
of COVID-19 screening systems, but traditional deep 
learning methods are usually computationally intensive.
To address the above problems, we propose MiniSeg, a lightweight
deep learning model for efficient COVID-19 segmentation.
Compared with traditional segmentation methods, MiniSeg has 
several significant strengths:
i) it only has 83K parameters and is thus not easy to overfit;
ii) it has high computational efficiency and is thus convenient
for practical deployment;
iii) it can be fast retrained by other users using their private 
COVID-19 data for further improving performance.
In addition, we build a comprehensive COVID-19 segmentation
benchmark for comparing MiniSeg to traditional methods.
\end{abstract}

\section{Introduction}
As one of the most severe pandemics in human 
history, \textit{coronavirus disease 2019} (COVID-19) threatens 
global health with thousands of newly infected patients every day.
Effective screening of infected patients is of high importance 
to the fight against COVID-19.
The gold standard for COVID-19 diagnosis is the tried-and-true 
Reverse Transcription Polymerase Chain Reaction (RT-PCR) testing 
\cite{wang2020detection}.
Unfortunately, the sensitivity of RT-PCR testing is not high 
enough to prevent the spread of COVID-19
\cite{ai2020correlation,fang2020sensitivity}.
Hence, computed tomography (CT) is used as a complementary tool
for RT-PCR testing to improve screening sensitivity
\cite{ai2020correlation,fang2020sensitivity}.
Besides, CT analysis is necessary for clinical monitoring of disease 
severity \cite{inui2020chest}.
However, CT examination needs expert radiologists, 
but we severely lack experienced radiologists during this pandemic.
Therefore, computer-aided systems are expected for automatic 
CT interpretation.

When it comes to computer-aided COVID-19 screening, 
deep-learning-based technology is a good choice due to its 
uncountable successful stories
\cite{sun2014deep,he2015delving,liu2019richer,liu2020rethinking,liu2018semantic,liu2018deep}.
However, directly applying traditional deep learning models 
for COVID-19 screening is suboptimal.
On the one hand, these models usually have millions of parameters
and thus require a large amount of labeled data for training.
The problem is that the publicly available COVID-19 data 
are limited and thus easy to cause overfitting for traditional
data-hungry models.
On the other hand, traditional deep learning methods, especially
the ones for image segmentation, are usually computationally
intensive.
Considering the current severe pandemic situation, fast 
training/testing and low computational load are essential 
for quick deployment and development of computer-aided 
COVID-19 screening systems.

It is a widely accepted concept that overfitting is easier to 
happen when a model has more parameters and less training data.
To solve the above problems of COVID-19 segmentation, we observe 
that lightweight networks are not only uneasy to overfit
owing to their small number of parameters but also likely to be 
efficient, making them suitable for computer-aided COVID-19 
screening systems.
Therefore, we think lightweight COVID-19 segmentation should
be the technical solution of this paper.
The key is to achieve accurate segmentation under the 
constraints of the number of network parameters and high
efficiency.
To achieve this goal, we find that the accuracy of image segmentation 
can be improved with effective multi-scale learning, which has 
significantly pushed forward the state of the arts of segmentation
\cite{chen2018deeplab,chen2017rethinking,chen2018encoder,yang2018denseaspp,zhao2017pyramid,mehta2018espnet,mehta2019espnetv2,pohlen2017full,yu2018learning}.
Hence, we resort to multi-scale learning 
to ensure the segmentation accuracy of lightweight networks.

With the above analyses, our effort starts with the design of an
\textbf{\underline{A}}ttentive 
\textbf{\underline{H}}ierarchical 
\textbf{\underline{S}}patial
\textbf{\underline{P}}yramid (\textbf{AHSP}) module for effective,
lightweight multi-scale learning.
AHSP first builds a spatial pyramid of dilated depthwise 
separable convolutions and feature pooling for learning 
multi-scale semantic features.
Then, the learned multi-scale features are fused hierarchically 
to enhance the capacity of multi-scale representation.
Finally, the multi-scale features are merged under the guidance 
of the attention mechanism, which learns to highlight essential 
information and filter out noisy information in radiography images.
With the AHSP module incorporated, we propose an extremely 
minimum network for efficient segmentation of COVID-19 
infected areas in chest CT slices.
Our model, namely \textbf{MiniSeg}, only has 83K parameters, 
two orders of magnitude less than traditional image segmentation 
methods, so that current limited COVID-19 data can be enough 
for training MiniSeg.
At last, we build a comprehensive COVID-19 segmentation benchmark
to compare MiniSeg to previous methods extensively.
Experiments demonstrate that MiniSeg performs favorably against
previous \sArt segmentation methods with high efficiency,
trained with limited COVID-19 data.

In summary, our contributions are threefold:
\begin{itemize}
\item We propose an \textbf{\underline{A}}ttentive 
\textbf{\underline{H}}ierarchical 
\textbf{\underline{S}}patial
\textbf{\underline{P}}yramid (\textbf{AHSP}) module for 
effective, lightweight multi-scale learning that is 
essential for image segmentation.
\item With AHSP incorporated, we present an extremely 
minimum network, \textbf{MiniSeg}, for accurate and efficient 
COVID-19 segmentation with limited training data.
\item For an extensive comparison of MiniSeg with previous \sArt
segmentation methods, we build a comprehensive COVID-19 
segmentation benchmark.
\end{itemize}

\section{Related Work}

\textbf{Image segmentation}
is a hot topic due to its wide range of applications.
Multi-scale learning plays an essential role in image segmentation
because objects in images usually exhibit very large scale changes.
Hence most current \sArt methods aim at designing
\textit{fully convolutional networks} (FCNs) \cite{shelhamer2017fully}
to learn effective multi-scale representations from input images. 
For example, U-Net \cite{ronneberger2015u}, 
U-Net++ \cite{zhou2018unet++},
and Attention U-Net \cite{oktay2018attention} 
propose encoder-decoder architectures to fuse multi-scale
deep features at multiple levels.
DeepLab \cite{chen2018deeplab} and its variants 
\cite{chen2017rethinking,chen2018encoder,yang2018denseaspp} 
design ASPP modules using dilated convolutions with different 
dilation rates to learn multi-scale features.
Besides the multi-scale learning, some studies focus on exploiting 
the global context information through pyramid pooling 
\cite{zhao2017pyramid}, context encoding \cite{zhang2018context},
or non-local operations \cite{huang2019ccnet,zhu2019asymmetric}.
The above models aim at improving segmentation accuracy 
without considering the model size and inference speed, 
so they are suboptimal for COVID-19 segmentation that 
only has limited training data and requires high efficiency.

\myPara{Lightweight networks}
aim at reducing the parameters and improving the efficiency 
of deep networks.
Convolutional factorization is an intuitive way to reduce the 
computational complexity of convolution operations.
Specifically, many well-known network architectures decompose the 
standard convolution into multiple steps to reduce the computational 
complexity, including 
Flattened Model \cite{jin2015flattened}, 
Inception networks \cite{szegedy2017inception}, 
Xception \cite{chollet2017xception}, 
%ResNeXt \cite{xie2017aggregated}, 
MobileNets \cite{howard2017mobilenets,sandler2018mobilenetv2},
and ShuffleNets \cite{zhang2018shufflenet,ma2018shufflenet}.
Among them, Xception and MobileNets factorize a 
convolution into a pointwise convolution 
and a depthwise separable convolution.
ShuffleNets further factorize a pointwise convolution into a channel 
shuffle operation and a grouped pointwise convolution.
There are also some studies focusing on efficient semantic 
segmentation network design
\cite{wu2018cgnet,mehta2018espnet,mehta2019espnetv2,lo2019efficient,wang2019lednet}.
Considering COVID-19 segmentation, our goal is to achieve higher 
accuracy and faster speed by enhancing multi-scale learning 
in a lightweight setting.

\myPara{Computer-aided COVID-19 screening}
has attracted much attention 
to serve as a supplementary tool for RT-PCR testing to improve
screening sensitivity.
Some studies
\cite{narin2020automatic,gozes2020rapid,xu2020deep,li2020artificial,zhang2020covid,wang2020covid}
design deep neural networks to classify chest X-rays or CT slices for 
COVID-19 screening.
\citet{fan2020inf} proposed a segmentation model for COVID-19 infected 
area segmentation from CT slices.
However, their method also falls into the same category as previous
segmentation methods and is thus suboptimal.
Some public COVID-19 imaging datasets, such as
COVID-19 X-ray Collection \cite{cohen2020covid}, 
COVID-CT-Dataset \cite{zhao2020covid},
COVID-19 CT Segmentation Dataset \cite{covid-data},
and COVID-19-CT-Seg \cite{jun2020covid},
are introduced.
In this paper, we focus on segmenting COVID-19 infected areas
from chest CT slices.

\section{Methodology} \label{ssec:method}

\subsection{Attentive Hierarchical Spatial Pyramid Module}
Although the factorization of a convolution operation into 
a pointwise convolution and a depthwise separable convolution
(\textbf{DSConv}) can significantly reduce the number of 
network parameters and computational complexity, it usually 
comes with the degradation of accuracy
\cite{howard2017mobilenets,sandler2018mobilenetv2,zhang2018shufflenet,ma2018shufflenet}.
Inspired by the fact that effective multi-scale learning 
plays an essential role in improving segmentation accuracy
\cite{chen2018deeplab,chen2017rethinking,chen2018encoder,yang2018denseaspp,zhao2017pyramid,mehta2018espnet,mehta2019espnetv2,pohlen2017full,yu2018learning},
we propose the AHSP module for effective and efficient multi-scale 
learning in a lightweight setting.
Besides some common convolution operations, such as vanilla
convolution, pointwise convolution, and DSConv, we introduce 
the dilated DSConv convolution that adopts a dilated 
convolution kernel for each input channel.
Suppose $\mathcal{F}_{r}^{k \times k}$ denotes a vanilla 
convolution, where $k \times k$ is the size of convolution
kernel and $r$ is the dilation rate.
Suppose $\hat{\mathcal{F}}_{r}^{k \times k}$ denotes a depthwise
separable convolution, where $k \times k$ and $r$ have the same
meaning as $\mathcal{F}_{r}^{k \times k}$.
The subscript $r$ will be omitted without ambiguity if we have 
a dilation rate of 1, \ie $r=1$.
For example, $\mathcal{F}^{1 \times 1}$ represents a pointwise
convolution (\ie $1 \times 1$ convolution).
$\hat{\mathcal{F}}_{2}^{3 \times 3}$ is a dilated 
$3 \times 3$ DSConv with a dilation rate of 2.

\begin{figure}[!tb]
\centering
\includegraphics[width=\linewidth]{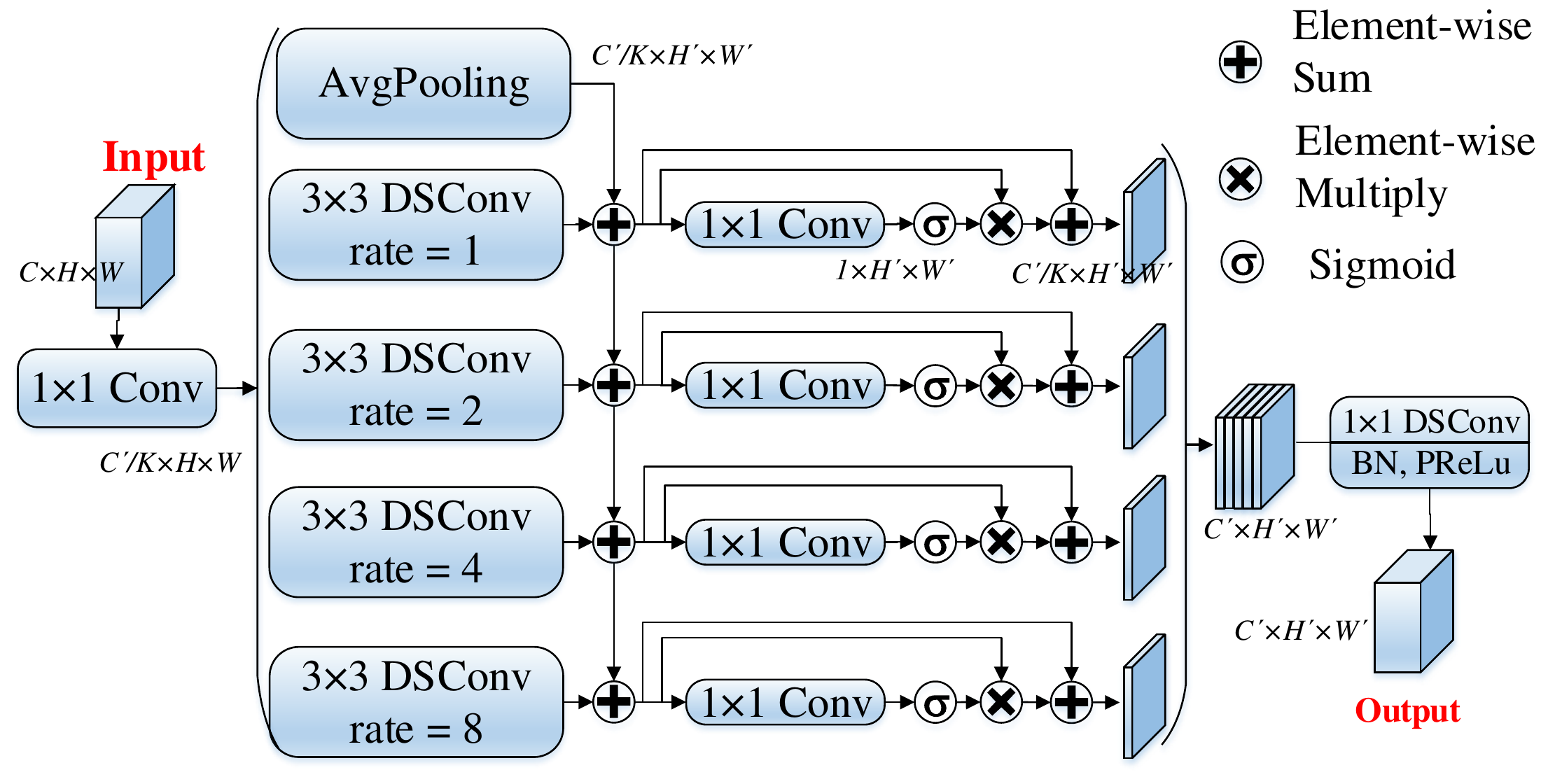}
\caption{Illustration of the proposed AHSP module.}
\label{fig:AHSP}
\vspace{-0.1in}
\end{figure}

With the above definitions for basic operations, we continue by 
introducing the proposed AHSP module illustrated in \figref{fig:AHSP}.
Let $\mathbf{X}\in\mathbb{R}^{C\times H\times W}$ be the 
input feature map so that the output feature map is 
$\mathcal{E}(\mathbf{X})\in\mathbb{R}^{C'\times H'\times W'}$,
where $\mathcal{E}$ denotes the transformation function
of AHSP for its input.
$C$, $H$, and $W$ are the number of channels, height, and 
width of the input feature map $\mathbf{X}$, respectively.
Similar definitions hold for $C'$, $H'$, and $W'$.
The input feature map $\mathbf{X}$ is first processed 
by a pointwise convolution to shrink the number of channels
into $C'/K$, in which $K$ is the number of parallel 
branches which will be described later.
This operation can be written as 
\begin{equation}\label{equ:shrink}
\mathbf{S}=\mathcal{F}^{1 \times 1}(\mathbf{X}).
\end{equation}
Then, the generated feature map $\mathbf{S}$ is fed into
$K$ parallel dilated DSConv, \ie
\begin{equation}\label{equ:dsconv}
\mathbf{F}_k = \hat{\mathcal{F}}_{2^{k-1}}^{3 \times 3}(\mathbf{S}),
\quad k = 1, 2, \cdots, K,
\end{equation}
where the dilation rate is increased exponentially for enlarging
the receptive field.
\equref{equ:dsconv} is the basis for multi-scale 
learning with large dilation rates capturing large-scale contextual
information and small dilation rates capturing local information.
We also add an average pooling operation for $\mathbf{S}$
to enrich the multi-scale information, \ie
\begin{equation}\label{equ:pool}
\mathbf{F}_0 = {\rm AvgPool}^{3\times 3}(\mathbf{S}),
\end{equation}
where ${\rm AvgPool}^{3\times 3}$ represents the average pooling
with a kernel size of $3\times 3$.
Note that we have 
$\mathbf{F}_k\in\mathbb{R}^{\frac{C'}{K}\times H'\times W'}$
for $k=0,1,\cdots,K$.
If we have $H\ne H'$ or $W\ne W'$, the convolution
and pooling operations in \equref{equ:dsconv} and 
\equref{equ:pool} will have a stride of 2 to downsample
the feature map by a scale of 2; otherwise, the stride 
will be 1.

These multi-scale feature maps produced by \equref{equ:dsconv}
and \equref{equ:pool} are aggregated in an attentive hierarchical
manner.
We first add them up hierarchically as 
\begin{equation}\label{equ:hierarchical}
\begin{aligned}
\dot{\mathbf{F}}_1 &= \mathbf{F}_0 + \mathbf{F}_1,\\
\dot{\mathbf{F}}_2 &= \dot{\mathbf{F}}_1 + \mathbf{F}_2,\\
&\cdots\\
\dot{\mathbf{F}}_K &= \dot{\mathbf{F}}_{K-1} + \mathbf{F}_K,
\end{aligned}
\end{equation}
where feature maps are gradually fused from small scales
to large scales to enhance the representation capability 
of multi-scale learning.
We further adopt a spatial attention mechanism to make 
the AHSP module automatically learn to focus on target 
structures of various scales.
On the other hand, the attention mechanism can also learn to
suppress irrelevant information at some feature scales and 
emphasize essential information at other scales.
Such self-attention makes each scale speak for itself to decide 
how important it is in the multi-scale learning process.
The transformation of $\dot{\mathbf{F}}$ by spatial attention
can be formulated as 
\begin{equation}\label{equ:attention}
\ddot{\mathbf{F}}_k = \dot{\mathbf{F}}_k + \dot{\mathbf{F}}_k \otimes \sigma(\mathcal{F}^{1 \times 1}(\dot{\mathbf{F}}_k)),
\quad k=1,2,\cdots,K,
\end{equation}
in which $\sigma$ is a \textit{sigmoid} activation function
and $\otimes$ indicates element-wise multiplication.
The pointwise convolution in \equref{equ:attention} outputs 
a single-channel feature map which is then transformed to 
a spatial attention map by the \textit{sigmoid} function.
This attention map is replicated to the same size as 
$\dot{\mathbf{F}}_k$, \ie $\frac{C'}{K}\times H'\times W'$,
before element-wise multiplication.
Considering the efficiency, we can compute the attention 
map for all $K$ branches together, like
\begin{equation}\label{equ:attention2}
\mathbf{A} = \sigma(\mathcal{F}^{1\times 1}({\rm Concat}(\dot{\mathbf{F}}_1, \dot{\mathbf{F}}_2, \cdots, \dot{\mathbf{F}}_K))),
\end{equation}
where ${\rm Concat}(\cdot)$ means to concatenate a series of feature
maps along the channel dimension.
The pointwise convolution in \equref{equ:attention2} is 
a $K$-grouped convolution with $K$ output channels, so we have
$\mathbf{A}\in\mathbb{R}^{K\times H'\times W'}$.
Hence, we can rewrite \equref{equ:attention} as 
\begin{equation}\label{equ:attention3}
\ddot{\mathbf{F}}_k = \dot{\mathbf{F}}_k + \dot{\mathbf{F}}_k \otimes \mathbf{A}[k], \quad k=1,2,\cdots,K,
\end{equation}
in which $\mathbf{A}[k]$ means the $k$-th channel of $\mathbf{A}$.

Finally, we merge and fuse the above hierarchical feature maps as
\begin{equation}\label{equ:fuse}
\begin{aligned}
\ddot{\mathbf{F}}\hspace{3mm} &= {\rm Concat}(\ddot{\mathbf{F}}_1, \ddot{\mathbf{F}}_2, \cdots, \ddot{\mathbf{F}}_K),\\
\mathcal{E}(\mathbf{X}) &= {\rm PReLU}({\rm BatchNorm}(\mathcal{F}^{1\times 1}(\ddot{\mathbf{F}}))),
\end{aligned}
\end{equation}
where ${\rm BatchNorm}(\cdot)$ denotes the batch
normalization \cite{ioffe2015batch} and ${\rm PReLU}(\cdot)$ 
indicates PReLU (\ie Parametric ReLU) activation function
\cite{he2015delving}.
The pointwise convolution in \equref{equ:fuse} is a $K$-grouped
convolution with $C'$ output channels, so that this pointwise
convolution aims at fusing $\ddot{\mathbf{F}}_k$ ($k=1,2,\cdots,K$)
separately, \ie adding connection to channels for depthwise 
convolutions in \equref{equ:dsconv}.
The fusion among various feature scales is achieved through the first
pointwise convolution (\ie \equref{equ:shrink}) in the subsequent 
AHSP module of MiniSeg
and the hierarchical feature aggregation (\ie \equref{equ:hierarchical}).
Such a design can reduce the number of convolution parameters 
in \equref{equ:fuse} by $K$ times when compared with that 
using a vanilla pointwise convolution, \ie $C'^2/K$ \vs $C'^2$.

Given an input feature map 
$\mathbf{X}\in\mathbb{R}^{C\times H\times W}$, we can compute 
the output feature map
$\mathcal{E}(\mathbf{X})\in\mathbb{R}^{C'\times H'\times W'}$
of an AHSP module using \equref{equ:shrink} - \equref{equ:fuse}.
We can easily find that increasing $K$ will reduce the number 
of AHSP parameters.
Considering the balance between segmentation accuracy and 
efficiency, we set $K=4$ in our experiments.
The proposed AHSP module not only significantly reduces the 
number of parameters but also enables us to learn effective 
multi-scale features so that we can adopt the limited COVID-19
data to train a high-quality segmenter.

\begin{figure}[!tb]
\centering
\includegraphics[width=\linewidth]{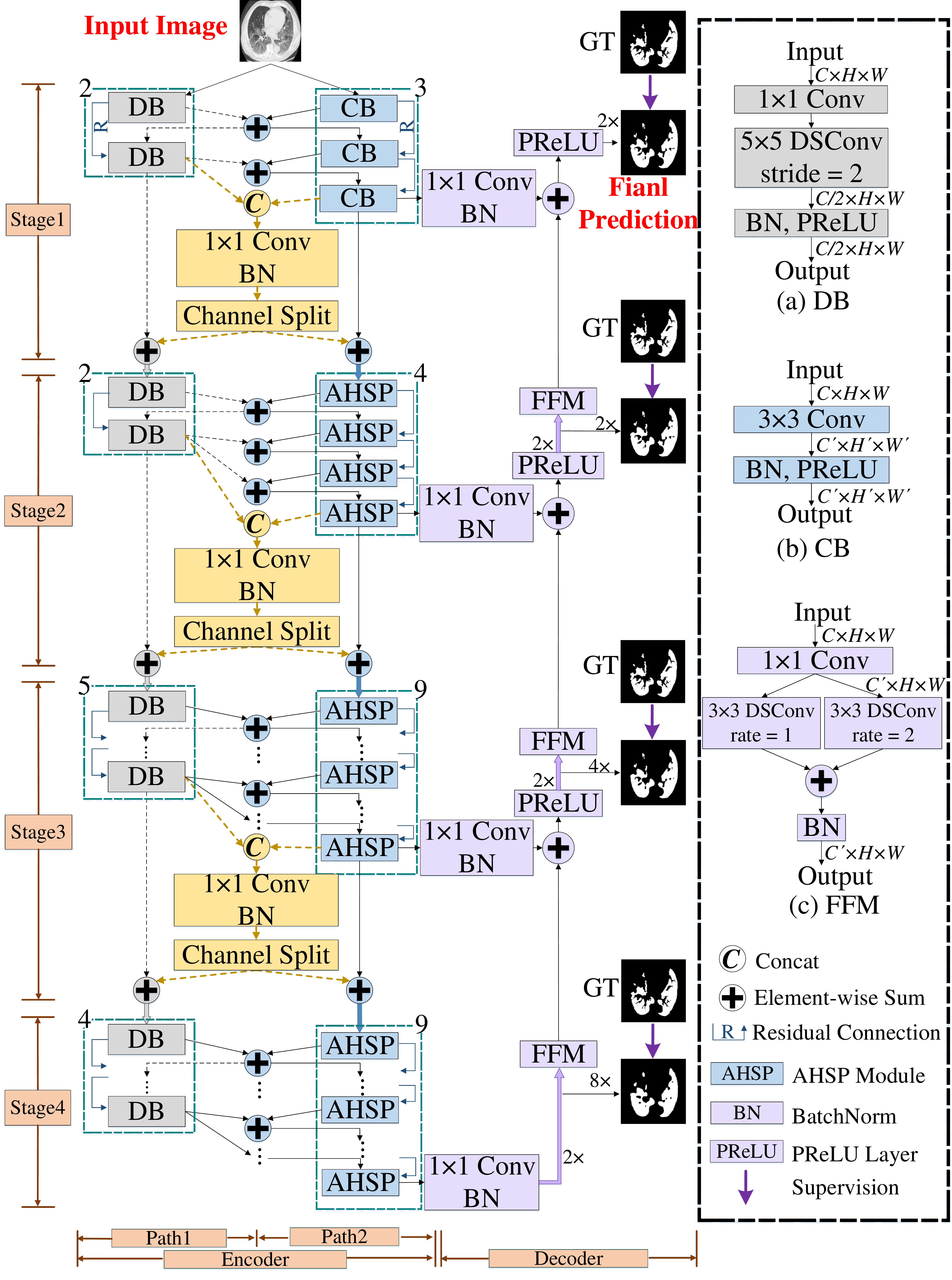}
\caption{Network architecture of the proposed MiniSeg.}
\label{fig:frame}
\vspace{-0.1in}
\end{figure}

\subsection{Network Architecture}
MiniSeg has an encoder-decoder structure.
The encoder sub-network focuses on learning effective multi-scale 
representations for the input image.
The decoder sub-network gradually aggregates the representations 
at different levels of the encoder to predict COVID-19 infected areas.
The network architecture of MiniSeg is displayed 
in \figref{fig:frame}.

\myPara{The encoder sub-network} uses AHSP as the basic module,
consisting of two paths connected through a series of nested skip pathways.
Suppose $\mathbf{I}\in\mathbb{R}^{3\times H\times W}$ denotes 
an input chest CT slice, where a grayscale CT slice is replicated 
three times to make its number of channels the same as color images.
The input $\mathbf{I}$ is downsampled four times,
resulting in four scales of $1/2$, $1/4$, $1/8$, and $1/16$,
with four stages processing such four scales, respectively.
Downsampling happens in the first block of each stage.

Suppose in the encoder sub-network we denote the output feature 
map of the $i$-th stage and the $j$-th block as $\mathbf{E}^i_j$, 
\textit{w.r.t.} $i\in\{1,2,3,4\}$ and $j\in\{1,2,\cdots,N_i\}$, 
where $N_i$ indicates the number of blocks at the $i$-th stage.
Therefore, we have 
$\mathbf{E}^i_j\in\mathbb{R}^{C_i\times\frac{H}{2^i}\times\frac{W}{2^i}}$,
in which $C_i$ is the number of feature channels 
at the $i$-th stage.
The abovementioned block refers to the proposed AHSP module 
except for the first stage whose basic block is the 
vanilla \textbf{Convolution Block} (\textbf{CB}).
Since the number of feature channels at the first stage (\ie 
$C_1$) is small, the vanilla convolution will not introduce 
too many parameters.
Without ambiguity, let $\mathcal{E}^i_j(\cdot)$ be the 
transformation function of the $i$-th stage and the $j$-th block
without distinguishing whether this block is a vanilla 
convolution or an AHSP module.
For the another path, we propose a \textbf{Downsampler Block} 
(\textbf{DB}).
The transformation function of a DB block is denoted as
$\mathcal{Q}^i_k(\cdot)$, \textit{w.r.t.} $i\in\{1,2,3,4\}$ and $k\in\{1,2,\cdots,M_i\}$, where $M_i$ denotes the number of DB
at the $i$-th stage.
We define DB as 
\begin{equation}
\mathcal{Q}^i_k(\mathbf{X}) = {\rm PReLU}({\rm BatchNorm}(\hat{\mathcal{F}}^{5\times 5}(\mathcal{F}^{1\times 1}(\mathbf{X})))),
\end{equation}
where $\hat{\mathcal{F}}^{5\times 5}(\cdot)$ has a stride 
of 2 for downsampling when we have $k=1$.
Suppose the output of $\mathcal{Q}^i_k(\cdot)$ is 
$\mathbf{Q}^i_k$.

Therefore, for the first block of the first stage, we have
\begin{equation}
\mathbf{E}^1_1 = \mathcal{E}^1_1(\mathbf{I}), \quad
\mathbf{Q}^1_1 = \mathcal{Q}^1_1(\mathbf{I}).
\end{equation}
For the first block of other stages, we compute the output 
feature map as
\begin{equation}\label{equ:concat-fuse-split}
\begin{aligned}
\mathbf{E}^{i-1} &= \mathcal{F}^{1 \times 1}({\rm Concat}(\mathbf{E}^{i-1}_{N_{i-1}}, \mathbf{Q}^{i-1}_{M_{i-1}})),\\
\mathbf{E}^i_1 &= \mathcal{E}^i_1({\rm Split}(\mathbf{E}^{i-1}) + \mathbf{E}^{i-1}_{N_{i-1}}),\\
\mathbf{Q}^i_1 &= \mathcal{Q}^i_1({\rm Split}(\mathbf{E}^{i-1}) + \mathbf{Q}^{i-1}_{M_{i-1}}),
\end{aligned}
\end{equation}
where we have $i\in \{2,3,4\}$.
The operation ${\rm Split}(\cdot)$ is to split a feature map along 
the channel dimension into two chunks, which are fed into
$\mathcal{E}^i_1$ and $\mathcal{Q}^i_1$, respectively.
Here, $\mathcal{E}^i_1(\cdot)$ and $\mathcal{Q}^i_1(\cdot)$
($i\in\{1,2,3,4\}$) have a stride of 2 for downsampling.
Instead of only using on-the-fly element-wise sum 
(\equref{equ:conn} and \equref{equ:sec_path}),
through \equref{equ:concat-fuse-split}, 
we conduct a ``concat-fuse-split'' operation 
to fully integrate the features from the two paths,
as concatenation can do better for feature fusion than sum 
by avoiding the information loss of sum \cite{huang2017densely}.
${\rm Split}(\cdot)$ is used to handle the increased number 
of channels brought by concatenation.

For other blocks, the output feature map is 
\begin{equation}\label{equ:conn}
\begin{gathered}
\mathbf{E}^i_j = \mathcal{E}^i_j(\mathbf{E}^i_{j-1} + \mathbf{Q}^i_{j'}) + \mathbf{E}^i_{j-1},\\ 
\textit{\text{w.r.t. }} i\in \{1,2,3,4\} \text{ and } j\in \{2,3,\cdots,N_i\},
\end{gathered}
\end{equation}
where $\mathcal{E}^i_j(\cdot)$ has a stride of 1 and a residual
connection is included for better optimization.
We have $j'=j-1$ if we also have $j - 1\leq M_i$; 
otherwise, we have $j'=M_i$.
The computation of $\mathbf{Q}^i_k$ can be formulated as 
\begin{equation}\label{equ:sec_path}
\begin{gathered}
\mathbf{Q}^i_k = \mathcal{Q}^i_k(\mathbf{Q}^i_{k-1} + \mathbf{E}^i_{k-1}) + \mathbf{Q}^i_{k-1},\\ 
\textit{\text{w.r.t. }} i\in \{1,2,3,4\} \text{ and } k\in \{2,3,\cdots,M_i\}.
\end{gathered}
\end{equation}
Through \equref{equ:conn} and \equref{equ:sec_path}, the two paths
of the encoder sub-network build nested skip connections.
Such a design benefits the multi-scale learning of the encoder.
Considering the balance among the number of network parameters,
segmentation accuracy, and efficiency, 
we set $C_i$ to $\{8, 24, 32, 64\}$, $N_i$ to $\{3, 4, 9, 9\}$, 
and $M_i$ to $\{2, 2, 5, 4\}$ for $i\in \{1, 2, 3, 4\}$, respectively.

\myPara{The decoder sub-network} is simple for efficient 
multi-scale feature decoding.
Since the top feature map of the encoder has a scale of $1/16$
of the original input, it is suboptimal to predict COVID-19
infected areas directly due to the loss of fine details.
Instead, we utilize a simple decoder sub-network to gradually
upsample and fuse the learned feature map at each scale.
A \textbf{Feature Fusion Module} (\textbf{FFM}) is proposed 
for feature aggregation.
Let $\mathcal{D}_i(\cdot)$ represent the function of FFM:
\begin{equation}
\begin{aligned}
\mathbf{S}'_i \hspace{3mm} &= \mathcal{F}^{1\times 1}(\mathbf{X}),\\
\mathcal{D}_i(\mathbf{X}) &= {\rm BatchNorm}(\hat{\mathcal{F}}^{3\times 3}(\mathbf{S}'_i) + \hat{\mathcal{F}}_2^{3\times 3}(\mathbf{S}'_i)), 
\end{aligned}
\end{equation}
in which $\mathcal{D}_i(\mathbf{X})$ ($i=1,2,3$) has 
$C_i$ channels as the pointwise convolution is utilized 
to adjust such number of channels.
We denote the feature map in the decoder as 
$\mathbf{D}_i \in \mathbb{R}^{C_i \times \frac{H}{2^i} \times \frac{W}{2^i}}$,
and we have $\mathbf{D}_4 = {\rm BatchNorm}(\mathcal{F}^{1\times 1}(\mathbf{E}_{N_4}^4))$.
We compute $\mathbf{D}_i$ ($i=3,2,1$) as
\begin{equation}
\begin{aligned}
\mathbf{S}''_i &= \mathcal{D}_i({\rm Upsample}(\mathbf{D}_{i+1},2)),\\
\mathbf{D}_i &= {\rm PReLU}(\mathbf{S}''_i + {\rm BatchNorm}(\mathcal{F}^{1\times 1}(\mathbf{E}_{N_i}^i))),
\end{aligned}
\end{equation}
where ${\rm Upsample}(\cdot,t)$ means to upsample 
a feature map by a scale of $t$ using bilinear interpolation.
In this way, the decoder sub-network enhances the high-level 
semantic features with low-level fine details, so that MiniSeg
can make accurate predictions for COVID-19 infected areas.

With $\mathbf{D}_i$ ($i=1,2,3,4$) computed, we can make dense 
prediction using a pointwise convolution, \ie
\begin{equation}\label{equ:pred}
\mathbf{P}_i = {\rm Softmax}({\rm Upsample}(\mathcal{F}^{1\times 1}(\mathbf{D}_i), 2^i)),
\end{equation}
where ${\rm Softmax}(\cdot)$ is the standard 
\textit{softmax} function and this pointwise convolution has 
two output channels representing two classes of background and 
COVID-19, respectively.
$\mathbf{P}_i \in \mathbb{R}^{H\times W}$ is the predicted 
class label map.
We utilize $\mathbf{P}_1$ as the final output prediction.
In training, we impose deep supervision \cite{lee2015deeply}
by replacing the \textit{softmax} function in \equref{equ:pred} 
with the standard cross-entropy loss function.

\begin{table}[!tb]
\centering
\setlength{\tabcolsep}{.8mm}
\resizebox{\linewidth}{!}{% ---> Don't forget this
\begin{tabular}{c|c|c} \Xhline{0.5pt}
    Datasets & \#Total/\#COVID & \#Patients
    \\ \hline
    COVID-19-CT100 \cite{covid-data} & 100/100 & $\sim$60 \\
    COVID-19-P9 \cite{covid-data} & 829/373 & 9 \\
    COVID-19-P20 \cite{jun2020covid} & 1844/1844 & 20 \\
    COVID-19-P1110 \cite{morozov2020mosmeddata} & 785/785 & 50
    \\ \Xhline{0.5pt}
    \end{tabular}}
    \caption{A summary of public COVID-19 CT datasets. \#Total and \#COVID 
    denote the numbers of all or COVID-19 infected CT slices, respectively.}
\label{tab:datasets}
\vspace{-0.1in}
\end{table}

\section{Experiments} \label{sec:experiments}
\subsection{Experimental Setup}
\subsubsection{Implementation details.}
We implement the proposed MiniSeg network using the 
well-known PyTorch framework \cite{paszke2017automatic}.
Adam optimization \cite{kingma2015adam} is used for training 
with the weight decay of 1e-4. 
We adopt the learning rate policy of \textit{poly}, where the initial 
learning rate is 1e-3.
We train 80 epochs on the training set with a batch size of 5.
We train all previous \sArt segmentation methods using 
the same training settings as MiniSeg for a fair comparison.

\subsubsection{Dataset.}
We utilize four open-access COVID-19 CT segmentation datasets, 
\ie two sub-datasets from \textit{COVID-19 CT Segmentation Dataset} \cite{covid-data},
\textit{COVID-19 CT Lung and Infection Segmentation Dataset} \cite{jun2020covid}, 
and \textit{MosMedData} \cite{morozov2020mosmeddata}, 
to evaluate MiniSeg.
According to the number of CT slices or the number of COVID-19 patients, 
we rename these datasets as COVID-19-CT100, COVID-19-P9,
COVID-19-P20, and COVID-19-P1110 for convenience, respectively.
The information of these datasets is summarized in \tabref{tab:datasets}.
We utilize the standard cropping and random flipping 
for data augmentation for MiniSeg and all baselines in training.
Moreover, we perform \textbf{5-fold cross-validation} to avoid
statistically significant differences in performance evaluation.

\begin{table}[!tb]
\centering
\setlength{\tabcolsep}{1.8mm}
\resizebox{\linewidth}{!}{% ---> Don't forget this
\begin{tabular}{c|c|c|c|c|ccccc} \Xhline{0.5pt}
    \multirow{2}{*}{SB} % Single Branch
    & \multirow{2}{*}{MB} % Multiple Branches
    & \multirow{2}{*}{AH} % Attentive Hierarchy
    & \multirow{2}{*}{TP} % Two Paths
    & \multirow{2}{*}{CS} % Channel Split
    & \multicolumn{5}{c}{Metrics (\%)} \\ \cline{6-10}
    & & & & & mIoU & SEN & SPE & DSC & HD 
    \\ \hline
    \ding{52} & & & & & 75.78 & 71.80 & 96.78 & 59.63 & 92.12
    \\
    & \ding{52} & & & & 76.31 & 76.22 & 97.40 & 61.57 & 88.05
    \\
    & \ding{52} & \ding{52} & & & 76.58 & 77.89 & 97.61 & 62.06 & 83.71
    \\
    & \ding{52} & \ding{52} & \ding{52} & & 76.66 & 78.72 & 97.02 & 62.05 & 78.67
    \\
    & \ding{52} & \ding{52} & \ding{52} & \ding{52} & 
    \textbf{78.33} & \textbf{79.62} & \textbf{97.71} & \textbf{64.84} & \textbf{71.69}
    \\ \Xhline{0.5pt}
\end{tabular}}
\caption{Effect of the main components in MiniSeg on the COVID-19-P1110 dataset. 
%A metric marked by $\uparrow$ means that a model is better 
%if it achieves higher results in terms of this metric, while 
%$\downarrow$ indicates the opposite.
Note that the metric HD does not have the unit of \%.}
\label{tab:ablation}
\end{table}

\begin{table}[!tb]
\centering
\renewcommand{\arraystretch}{0.82}
\setlength{\tabcolsep}{.8mm}
\resizebox{\linewidth}{!}{% ---> Don't forget this
\begin{tabular}{c|c|c|c|c|c|ccccc} \Xhline{0.5pt}
    \multirow{2}{*}{PReLU}
    & \multirow{2}{*}{DE} 
    & \multirow{2}{*}{DS} % Deep Supervision
    & \multirow{2}{*}{CB} % CB in 1st Stage
    & \multirow{2}{*}{\tabincell{c}{DB\\ $5\times 5$}} % $5\times 5$ DSConv in DB
    & \multirow{2}{*}{FFM} % FFM in Decoder
    & \multicolumn{5}{c}{Metrics (\%)} 
    \\ \cline{7-11}
    & & & & & & mIoU & SEN & SPE & DSC & HD
    \\ \hline
    ReLU & & & & & & 76.92 & 75.41 & 96.90 & 62.11 & 78.39
    \\
    & \ding{55} & & & & & 73.11 & 76.31 & 97.45 & 55.35 & 76.72
    \\
    & & \ding{55} & & & & 76.45 & \textbf{80.94} & 96.38 & 62.46 & 87.27
    \\
    & & & AHSP & & & 76.71 & 78.38 & 96.38 & 62.05 & 78.99
    \\
    & & & & $3\times 3$ & & 76.54 & 78.43 & 97.09 & 61.81 & 80.61
    \\
    & & & & & AHSP & 77.15 & 78.69 & 97.33 & 62.46 & 82.98
    \\
    & & & & & & \textbf{78.33} & 79.62 & \textbf{97.71} & \textbf{64.84} & \textbf{71.69}
    \\ \Xhline{0.5pt}
\end{tabular}}
\caption{Effect of some design choices on COVID-19-P1110. 
Each design choice is replaced with the operation in the 
table or directly removed (\ding{55}).
DE: Decoder. DS: Deep Supervision.
}
\label{tab:choice}
\vspace{-0.1in}
\end{table}

\subsubsection{Evaluation metrics.}
We evaluate COVID-19 segmentation accuracy 
using five popular evaluation metrics in medical imaging 
analysis, \ie mean intersection over union (mIoU), 
sensitivity (SEN), specificity (SPC), Dice similarity coefficient 
(DSC), and Hausdorff distance (HD).
Specifically, mIoU, SEN, SPC, and DSC range between 0 and 1.
The larger these values, the better the model.
Note that a lower value of HD indicates better segmentation 
accuracy.
Moreover, we also report the number of parameters, the number of FLOPs, 
and speed, tested using a $512\times 512$ input image and a TITAN RTX GPU.

\subsection{Ablation Studies}
\subsubsection{Effect of main components.}
As shown in \tabref{tab:ablation}, we start with 
a \textbf{single-branch (SB)} module that only has the DSConv 
with a dilation rate of 1.
We replace all AHSP modules in MiniSeg with such SB modules and 
remove the two-path design of the MiniSeg encoder
(the $1^\text{st}$ line of \tabref{tab:ablation}).
Then, we extend such an SB module to a \textbf{multi-branch (MB)}
module using the spatial pyramid as in the AHSP module
to demonstrates the importance of multi-scale learning
(the $2^\text{nd}$ line of \tabref{tab:ablation}).
Next, we add the \textbf{attentive hierarchical fusion strategy (AH)}
to get the AHSP module to proves the superiority of the 
attentive hierarchical fusion
(the $3^\text{rd}$ line of \tabref{tab:ablation}).
We continue by adding the \textbf{two-path design (TP)} to the 
encoder sub-network to validates that such a two-path design 
can benefit the network optimization
(the $4^\text{th}$ line of \tabref{tab:ablation}).
At last, we add the \textbf{channel split (CS)} operation to 
obtain the final MiniSeg model
(the $5^\text{th}$ line of \tabref{tab:ablation}).
These ablation studies demonstrate that the main components
in MiniSeg are all effective for COVID-19 segmentation.

\subsubsection{Effect of some design choices.}
We continue by evaluating the design choices of MiniSeg.
The results are provided in \tabref{tab:choice}.
First, we replace the PReLU activation function
with the ReLU function.
Second, we remove the decoder sub-network and change the stride of
the last stage from 2 to 1, so we can directly make predictions
at the scale of $1/8$ and upsample to the original size, as in
previous studies 
\cite{mehta2018espnet,lo2019efficient,wu2018cgnet,paszke2016enet,chen2018deeplab}.
Third, we remove deep supervision in training. 
Fourth, we replace Convolution Blocks (CB) in the 
first stage with AHSP modules.
Fifth, we replace the $5\times 5$ DSConv in the 
Downsampler Blocks (DB) with $3\times 3$ DSConv.
Sixth, we replace the Feature Fusion Modules (FFM) in the
decoder sub-network with AHSP modules.
The default setting achieves the best overall performance, 
demonstrating the effectiveness of our designs.

\begin{table}[!tb]
\centering
\renewcommand{\arraystretch}{0.82}
\setlength{\tabcolsep}{1.2mm}
\resizebox{\linewidth}{!}{% ---> Don't forget this
\begin{tabular}{lccrrr} \Xhline{0.5pt}
    Method & Backbone & ImageNet & \#Param
    & FLOPs & Speed \\ \hline
    U-Net & - & No & 8.43M & 65.73G & 57.3fps
    \\ \rowcolor{mycolor}
    FCN-8s & VGG16 & Yes & 15.53M & 105.97G & 4.5fps
    \\
    SegNet & - & No & 28.75M & 160.44G & 3.0fps
    \\ \rowcolor{mycolor} 
    FRRN & - & No & 17.30M & 237.70G & 15.8fps
    \\
    PSPNet & ResNet50 & Yes & 64.03M & 257.79G & 17.1fps
    \\ \rowcolor{mycolor}
    DeepLabv3 & ResNet50 & Yes & 38.71M & 163.83G & 25.3fps
    \\ 
    DenseASPP & - & No & 27.93M & 122.28G & 19.3fps 
    \\ \rowcolor{mycolor}
    DFN & ResNet50 & Yes & 43.53M & 81.88G & 56.2fps
    \\
    EncNet & ResNet50 & Yes & 51.25M & 217.46G & 18.1fps 
    \\ \rowcolor{mycolor} 
    DeepLabv3+ & Xception & Yes & 53.33M & 82.87G & 3.4fps
    \\
    BiSeNet & ResNet18 & Yes & 12.50M & 13.01G & 172.4fps
    \\ \rowcolor{mycolor}
    UNet++ & - & No & 8.95M & 138.37G & 26.8fps
    \\
    Attention U-Net & - & No & 8.52M & 67.14G & 49.2fps
    \\ \rowcolor{mycolor}
    OCNet & ResNet50 & Yes & 51.60M &  220.69G & 19.3fps
    \\
    DUpsampling & ResNet50 & Yes & 28.46M & 123.01G & 36.5fps
    \\  \rowcolor{mycolor}
    DANet & ResNet50 & Yes & 64.87M &  275.72G & 16.4fps
    \\
    CCNet & ResNet50 & Yes & 46.32M & 197.92G & 40.0fps
    \\ \rowcolor{mycolor} 
    ANNNet & ResNet50 & Yes & 47.42M & 203.07G & 32.8fps
    \\ 
    GFF & ResNet50 & Yes & 90.57M & 374.03G & 17.5fps
    \\ \rowcolor{mycolor} 
    Inf-Net & ResNet50 & Yes & 30.19M & 27.30G & 155.9fps
    \\ \hline
    MobileNet & MobileNet & Yes & 3.13M & 3.02G & 416.7fps
    \\  \rowcolor{mycolor}
    MobileNetv2 & MobileNetv2& Yes & 2.17M & 1.60G & 137.0fps
    \\ 
    ShuffleNet & ShuffleNet & Yes & 0.92M & 0.75G & 116.3fps
    \\ \rowcolor{mycolor}
    ShuffleNetv2 & ShuffleNetv2 & Yes & 1.22M & 0.77G & 142.9fps
    \\ 
    EfficientNet & EfficientNet & No & 8.37M & 13.19G & 48.1fps 
    \\ \hline \rowcolor{mycolor}
    ENet & - & No & 0.36M & 1.92G & 71.4fps
    \\ 
    ESPNet & - & No & 0.35M & 1.76G & 125.0fps
    \\ \rowcolor{mycolor}
    CGNet & - & No & 0.49M &  3.40G & 73.0fps
    \\ 
    ESPNetv2 & - & No & 0.34M & 0.77G & 73.0fps
    \\ \rowcolor{mycolor}
    EDANet & - & No & 0.68M & 4.43G & 147.1fps
    \\ 
    LEDNet & - & No & 2.26M & 6.32G & 94.3fps
    \\ \hline \rowcolor{mycolor}
    MiniSeg & - & No & \textbf{82.91K} & \textbf{0.50G} & \textbf{516.3fps}
    \\ 
    \Xhline{0.5pt}
\end{tabular}}
\caption{Comparison of MiniSeg to previous \sArt 
methods in terms of parameters, FLOPs, and speed.
}
\label{tab:compParameters}
\vspace{-0.1in}
\end{table}

\begin{figure}[!tb]
\centering
\subfloat[]{\includegraphics[width=.48\linewidth, height=.45\linewidth]{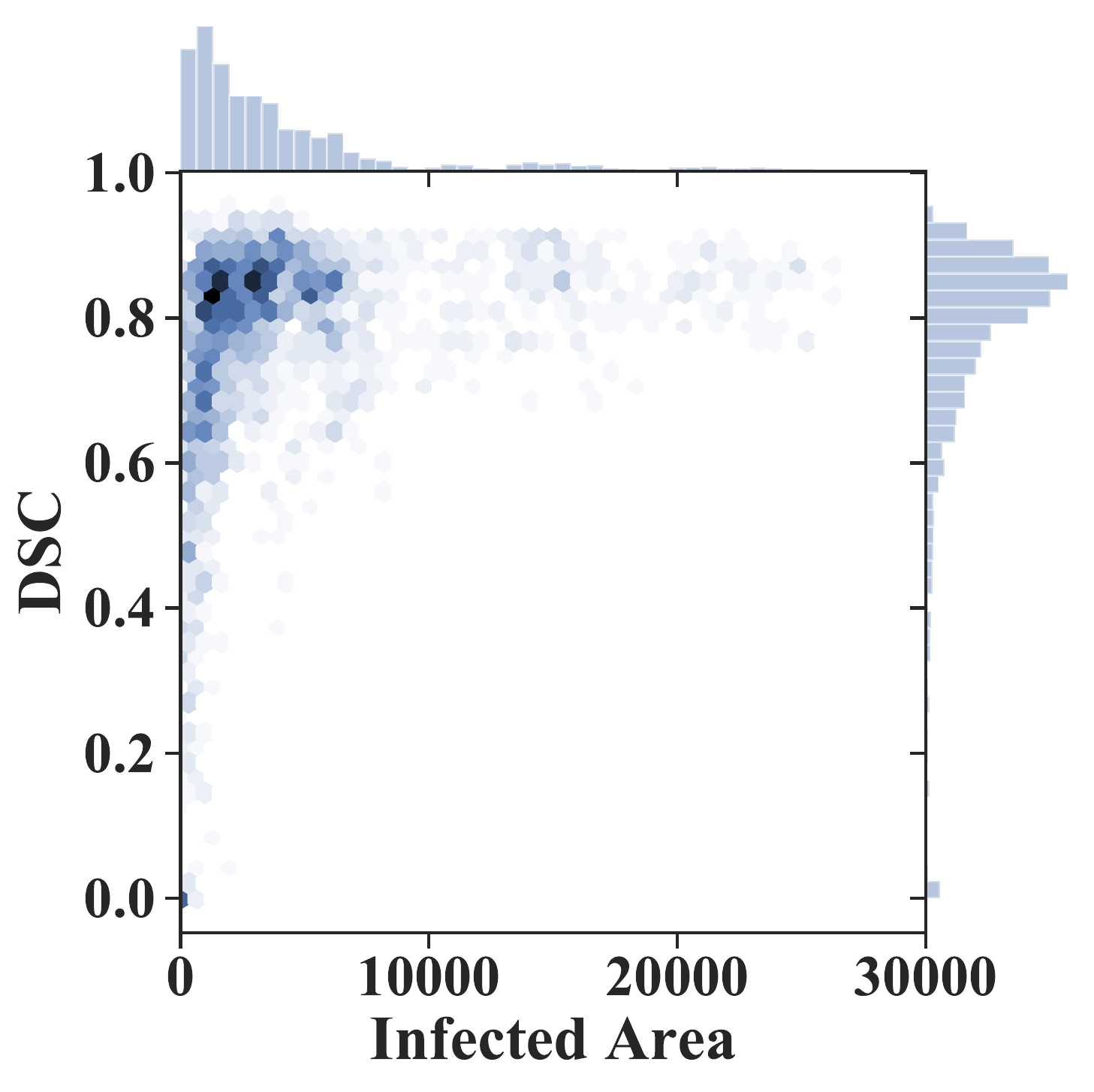}\label{fig:analysis_A}}
\subfloat[]{\includegraphics[width=.48\linewidth, height=.4\linewidth]{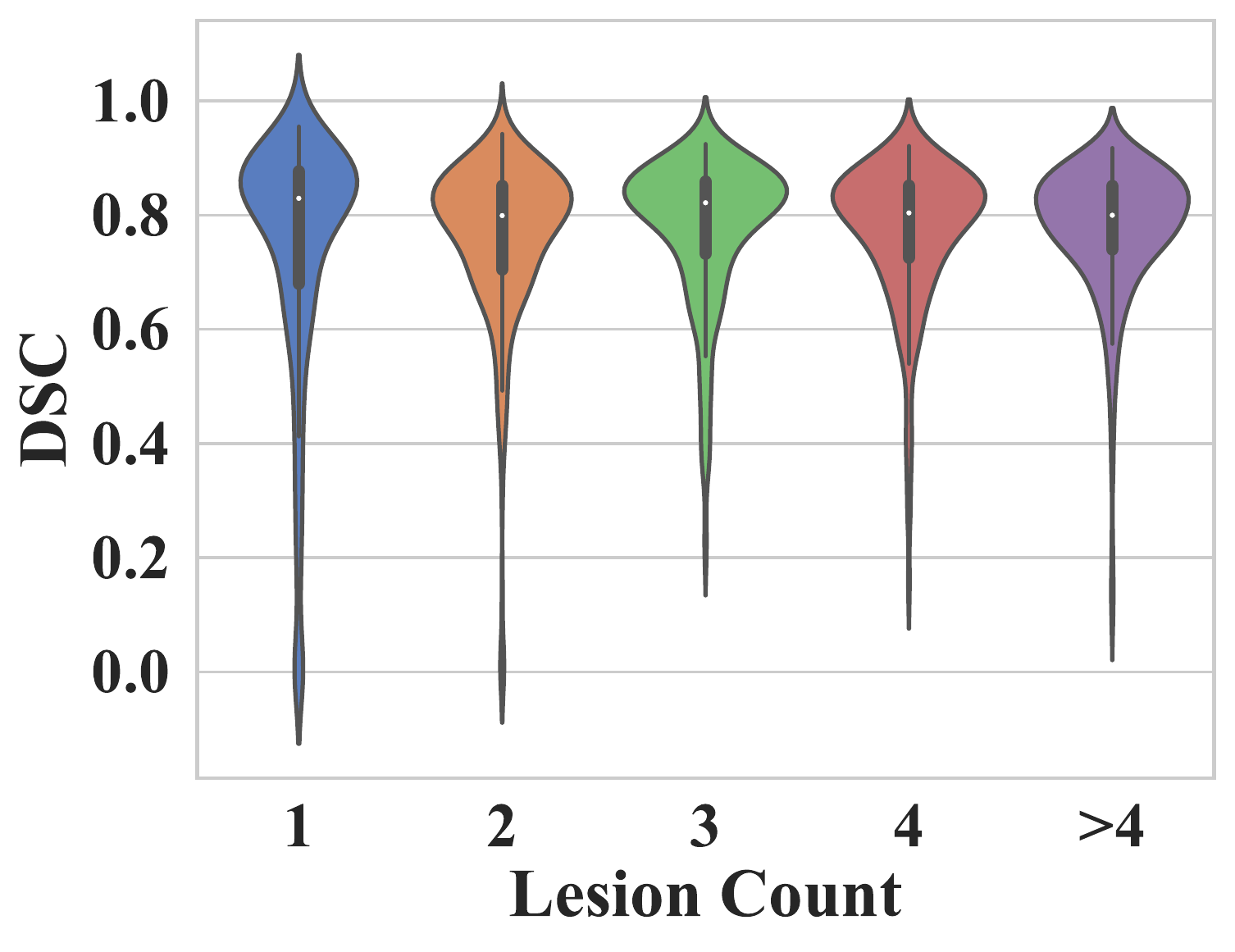}\label{fig:analysis_B}}
\\
\caption{Statistical analysis for MiniSeg on COVID-19-P20.
(a) The DSC score \vs the infected area;
(b) The DSC score \vs the lesion count in the corresponding CT slice.}
\vspace{-0.1in}
\end{figure}

\begin{table*}[!tb]
\centering
\renewcommand{\arraystretch}{0.90}
\setlength{\tabcolsep}{.8mm}
\resizebox{\linewidth}{!}{% ---> Don't forget this
\begin{tabular}{l|ccccr|ccccr|ccccr|ccccr} \Xhline{0.5pt}
    \multirow{2}{*}{Methods} 
    & \multicolumn{5}{c|}{COVID-19-CT100}
    & \multicolumn{5}{c|}{COVID-19-P9} 
    & \multicolumn{5}{c|}{COVID-19-P20} 
    & \multicolumn{5}{c}{COVID-19-P1110} 
    \\ \cline{2-21}
    & mIoU & SEN & SPC & DSC & HD  
    & mIoU & SEN & SPC & DSC & HD 
    & mIoU & SEN & SPC & DSC & HD 
    & mIoU & SEN & SPC & DSC & HD \\ \hline
    U-Net & 77.56 & 72.24 & 97.71 & 68.37 & 94.25
    & 76.51 & 88.53 & 98.93 & 65.69 & 133.64
    & 81.81 & 82.73 & 97.92 & 72.66 & 61.66
    & 74.26 & \textbf{81.85} & 97.33 & 58.62 & 95.72
    \\ \rowcolor{mycolor}
    FCN-8s & 71.85 & 66.47 & 93.56 & 58.11 & 104.68
    & 81.20 & 87.12 & 98.40 & 72.67 & 91.32
    & 82.54 & 84.10 & 98.02 & 73.60 & 51.47
    & 70.51 & 80.75 & 97.08 & 53.33 & 84.43
    \\
    SegNet & 75.02 & 80.02 & 96.34 & 64.84 & 109.05
    & 73.88 & 73.59 & 98.79 & 62.07 & 98.38
    & 79.55 & 81.68 & 98.44 & 69.68 & 77.28
    & 72.32 & 76.77 & 97.24 & 55.92 & 105.42
    \\ \rowcolor{mycolor}
    FRRN & 79.20 & 78.47 & 97.50 & 71.27 & 86.56
    & 80.83 & 86.26 & 99.54 & 74.03 & 84.34
    & 80.61 & 80.75 & 97.53 & 71.43 & 61.28
    & 73.84 & 75.45 & 95.80 & 58.86 & 87.11
    \\
    PSPNet & 75.61 & 70.82 & 96.47 & 64.55 & 99.76
    & 82.15 & 86.84 & 99.19 & 74.85 & 94.40
    & 81.60 & 83.44 & 98.17 & 71.60 & 65.60
    & 71.41 & 80.34 & 97.40 & 54.82 & 87.06
    \\ \rowcolor{mycolor} 
    DeepLabv3 & 81.30 & 84.80 & 97.48 & 74.65 & 81.77
    & 81.50 & 85.23 & 98.56 & 73.10 & 95.72
    & 80.26 & 81.60 & 97.78 & 70.96 & 60.50
    & 72.91 & 80.45 & 96.85 & 55.70 & 81.35
    \\
    DenseASPP & 78.43 & 81.14 & 97.02 & 70.37 & 156.23
    & 72.78 & 70.26 & 98.65 & 65.53 & 98.61
    & 81.11 & 82.21 & 97.80 & 71.68 & 64.05
    & 74.84 & 69.38 & 95.65 & 57.24 & 76.61
    \\ \rowcolor{mycolor}
    DFN & 81.07 & 84.27 & 97.49 & 74.45 & 83.73
    & 79.19 & 85.78 & 98.64 & 69.93 & 106.23
    & 79.13 & 80.96 & 96.51 & 69.46 & 66.56
    & 73.40 & 80.12 & 97.13 & 57.31 & 87.10
    \\
    EncNet & 71.28 & 74.11 & 95.20 & 62.83 & 119.55
    & 81.35 & 86.88 & 98.65 & 72.62 & 94.77
    & 82.43 & 84.94 & 98.03 & 71.60 & 71.57
    & 71.65 & 81.23 & 96.65 & 54.89 & 77.82
    \\ \rowcolor{mycolor}
    DeepLabv3+ & 79.45 & 79.58 & 97.55 & 71.70 & 93.09
    & 81.29 & 77.93 & 99.30 & 73.48 & 81.95
    & 81.26 & 81.61 & 95.35 & 42.79 & 182.14
    & 74.14 & 74.65 & 97.26 & 57.16 & 102.78
    \\
    BiSeNet & 63.09 & 74.07 & 87.41 & 58.66 & 110.47
    & 72.33 & 67.17 & 96.35 & 55.40 & 164.07
    & 78.08 & 76.13 & 97.07 & 65.24 & 85.94
    & 70.29 & 70.90 & 95.49 & 52.26 & 95.11
    \\ \rowcolor{mycolor}
    UNet++ & 77.64 & 77.26 & 97.28 & 69.04 & 91.73
    & 77.95 & 86.83 & 99.39 & 69.27 & 104.83
    & 80.73 & 79.61 & 96.75 & 70.34 & 63.01
    & 73.39 & 75.67 & 96.13 & 59.08 & 88.21
    \\
    Attention U-Net & 77.71 & 74.75 & 97.56 & 68.93 & 92.15
    & 76.26 & 76.39 & 99.24 & 66.74 & 102.43
    & 80.70 & 82.92 & 97.41 & 71.27 & 64.91
    & 74.62 & 81.32 & 97.63 & 59.34 & 95.16
    \\ \rowcolor{mycolor}
    OCNet & 69.29 & 72.86 & 89.38 & 56.14 & 105.66
    & 81.14 & 87.41 & 98.71 & 72.94 & 113.21
    & 80.74 & 80.71 & 95.82 & 69.36 & 56.60
    & 72.05 & 79.67 & 97.64 & 53.97 & 97.38
    \\
    DUpsampling & 81.69 & 84.54 & 97.60 & 75.27 & 81.07
    & 79.96 & 74.42 & 96.38 & 69.60 & 64.62
    & 81.05 & 79.37 & 96.34 & 71.01 & 60.19
    & 72.16 & 65.18 & 91.77 & 53.98 & 72.29
    \\ \rowcolor{mycolor}
    DANet & 73.57 & 66.30 & 92.76 & 61.34 & 99.11
    & 81.59 & 88.78 & 99.13 & 73.82 & 114.69
    & 78.35 & 79.87 & 97.31 & 67.04 & 83.13
    & 73.47 & 75.00 & 95.80 & 56.07 & 74.04
    \\ 
    CCNet & 75.24 & 69.55 & 95.92 & 63.99 & 98.03
    & 81.27 & 86.61 & 99.16 & 73.93 & 90.84
    & 82.22 & 82.93 & 97.76 & 73.13 & 56.98
    & 72.02 & 79.16 & 96.29 & 54.83 & 83.07
    \\ \rowcolor{mycolor}
    ANNNet & 73.93 & 66.73 & 95.72 & 62.06 & 102.43
    & 79.52 & 85.20 & 98.35 & 69.55 & 109.31
    & 81.92 & 84.10 & 98.13 & 72.72 & 56.99
    & 72.28 & 81.19 & 97.30 & 55.21 & 83.16
    \\ 
    GFF & 75.75 & 69.80 & 97.53 & 63.88 & 103.87
    & 81.20 & 85.35 & 98.46 & 72.61 & 113.48
    & 82.44 & 84.29 & 97.49 & 73.05 & 63.84
    & 71.82 & 81.10 & 96.50 & 53.88 & 86.39
    \\  \rowcolor{mycolor}
    Inf-Net & 81.62 & 76.50 & \textbf{98.32} & 74.44 & 86.81
    & 80.28 & 77.59 & 98.72 & 71.76 & 69.46
    & 64.62 & 69.46 & 99.02 & 63.38 & 79.68
    & 74.32 & 62.93 & 93.45 & 56.39 & 71.77
    \\ \hline
    MobileNet & 80.07 & 81.19 & 95.92 & 63.99 & 98.03
    & 81.32 & 85.53 & \textbf{99.62} & 74.18 & 128.95
    & 80.52 & 82.66 & 97.95 & 72.05 & 70.70
    & 74.84 & 80.08 & 97.67 & 59.91 & 92.88
    \\ \rowcolor{mycolor}
    MobileNetv2 & 79.73 & 82.83 & 97.32 & 72.53 & 88.40
    & 80.09 & 81.77 & 99.45 & 72.16 & 85.15
    & 80.99 & 83.16 & 98.20 & 71.50 & 68.54
    & 74.32 & 80.41 & 96.96 & 59.43 & 93.11
    \\ 
    ShuffleNet & 77.50 & 74.57 & 97.64 & 69.02 & 86.97
    & 80.87 & 83.62 & 99.28 & 72.66 & 105.56
    & 81.97 & 82.34 & 98.03 & 73.33 & 56.68 
    & 74.51 & 77.73 & 96.38 & 58.64 & 78.16
    \\ \rowcolor{mycolor}
    ShuffleNetv2 & 78.58 & 81.21 & 97.30 & 71.37 & 84.72
    & 79.54 & 82.44 & 98.75 & 70.29 & 102.75
    & 81.31 & 81.86 & 98.29 & 71.67 & 70.06
    & 74.56 & 76.89 & 96.58 & 58.67 & 78.55
    \\ 
    EfficientNet & 78.22 & 80.25 & 97.04 & 70.45 & 75.26
    & 73.13 & 73.50 & 99.25 & 60.20 & 133.45
    & 81.58 & 80.10 & 98.06 & 72.12 & 64.30
    & 73.30 & 80.66 & 97.07 & 58.04 & 96.30
    \\ \hline \rowcolor{mycolor}
    ENet & 79.49 & 81.26 & 97.53 & 71.57 & 96.08
    & 79.27 & 79.62 & 99.07 & 70.43 & 101.92
    & 77.57 & 76.35 & 97.16 & 68.23 & 67.40
    & 74.49 & 74.86 & 96.38 & 57.20 & 85.32
    \\ 
    ESPNet & 77.45 & 84.18 & 96.48 & 69.30 & 97.04
    & 76.79 & 71.30 & 98.67 & 67.68 & 93.58
    & 80.32 & 80.53 & 97.52 & 69.36 & 91.84
    & 74.75 & 72.06 & 96.96 & 57.77 & 94.58
    \\ \rowcolor{mycolor}
    CGNet & 79.34 & 81.55 & 96.34 & 71.42 & 90.37
    & 75.10 & 70.27 & 92.57 & 60.37 & 134.43
    & 82.24 & 80.73 & 97.35 & 72.35 & 53.63
    & 74.12 & 74.83 & 96.16 & 56.45 & 74.34
    \\ 
    ESPNetv2 & 78.66 & 77.84 & 96.53 & 70.46 & 87.77
    & 78.22 & 72.42 & 97.23 & 67.12 & 88.58
    & 80.78 & 79.03 & 97.41 & 70.13 & 73.67
    & 74.10 & 76.60 & 97.67 & 58.37 & 96.73
    \\ \rowcolor{mycolor}
    EDANet & 78.74 & 82.86 & 96.98 & 70.67 & 88.14
    & 80.11 & 79.40 & 98.77 & 72.89 & 70.40
    & 79.56 & 76.79 & 97.42 & 68.71 & 70.72
    & 73.21 & 73.73 & 96.71 & 55.11 & 84.56
    \\ 
    LEDNet & 77.41 & 81.69 & 96.93 & 68.74 & 92.49
    & 78.46 & 80.96 & 98.47 & 70.41 & 120.74
    & 80.34 & 78.74 & 97.90 & 70.10 & 65.77
    & 73.46 & 72.27 & 95.14 & 55.09 & 94.19
    \\ \hline \rowcolor{mycolor}
    MiniSeg 
    & \textbf{82.15} & \textbf{84.95} & 97.72 & \textbf{75.91} & \textbf{74.42}
    & \textbf{85.31} & \textbf{90.60} & 99.15 & \textbf{80.06} & \textbf{58.46}
    & \textbf{84.49} & \textbf{85.06} & \textbf{99.05} & \textbf{76.27} & \textbf{51.06}
    & \textbf{78.33} & 79.62 & \textbf{97.71} & \textbf{64.84} & \textbf{71.69}
    \\
    \Xhline{0.5pt}
\end{tabular}} \vspace{-0.05in}
\caption{Comparison between MiniSeg and previous \sArt 
segmentation methods. 
}
\label{tab:comp}
\vspace{-0.10in}
\end{table*}

\begin{figure*}[!t]
\centering
\includegraphics[width=\linewidth]{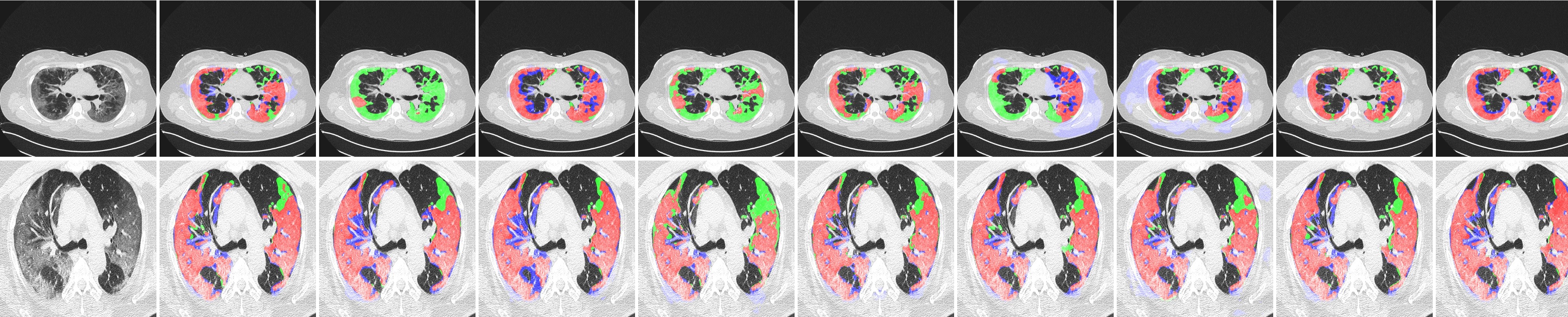}
\\ \vspace{-0.05in}
\leftline{\scriptsize\hspace{0.165in} CT Slice \hspace{0.35in} U-Net
    \hspace{0.43in} FCN \hspace{0.29in} DeepLabV3+ \hspace{0.24in} 
    UNet++ \hspace{0.18in} Attention UNet \hspace{0.21in} Inf-Net
    \hspace{0.25in} ShuffleNetV2 \hspace{0.2in} ESPNetv2 
    \hspace{0.27in} MiniSeg
} \vspace{-0.05in}
\caption{Visual comparison between MiniSeg and other methods.
Red: true positive; Green: false negative; Blue: false positive.}
\label{fig:cmp_sample}
\vspace{-0.10in}
\end{figure*}

\subsection{Comparison with State-of-the-art Methods}
\subsubsection{Quantitative Evaluation.}
To compare MiniSeg to previous \sArt competitors and 
promote COVID-19 segmentation research,
we build a comprehensive benchmark.
This benchmark contains 31 previous \sArt image segmentation 
methods, including 
U-Net \cite{ronneberger2015u}, 
FCN-8s \cite{shelhamer2017fully}, 
SegNet \cite{badrinarayanan2017segnet}, 
FRRN \cite{pohlen2017full}, 
PSPNet \cite{zhao2017pyramid}, 
DeepLabv3 \cite{chen2017rethinking}, 
DenseASPP \cite{yang2018denseaspp},
DFN \cite{yu2018learning}, 
EncNet \cite{zhang2018context}, 
DeepLabv3+ \cite{chen2018encoder},
BiSeNet \cite{yu2018bisenet}, 
UNet++ \cite{zhou2018unet++}, 
Attention U-Net \cite{oktay2018attention},
OCNet \cite{yuan2018ocnet},
DUpsampling \cite{tian2019decoders}, 
DANet \cite{fu2019dual},
CCNet \cite{huang2019ccnet},
ANNNet \cite{zhu2019asymmetric},
GFF \cite{li2020gated},
Inf-Net \cite{fan2020inf},
MobileNet \cite{howard2017mobilenets}, 
MobileNetv2 \cite{sandler2018mobilenetv2},
ShuffleNet \cite{zhang2018shufflenet},
ShuffleNetv2 \cite{ma2018shufflenet}, 
EfficientNet \cite{tan2019efficientnet},
ENet \cite{paszke2016enet},
ESPNet \cite{mehta2018espnet},
CGNet \cite{wu2018cgnet},
ESPNetv2 \cite{mehta2019espnetv2},
EDANet \cite{lo2019efficient},
and LEDNet \cite{wang2019lednet}.
Among them, Inf-Net is designed for COVID-19 segmentation.
MobileNet, MobileNetv2, ShuffleNet,
ShuffleNetv2, and EfficientNet are designed for
lightweight image classification.
We view them as the encoder and add the decoder of MiniSeg 
to them so that they are reformed as image segmentation models.
ENet, ESPNet, CGNet, ESPNetv2, EDANet, and LEDNet 
are well-known lightweight segmentation models.
The code of these methods is provided online by the authors.
We believe that this benchmark would be useful for future research 
on COVID-19 segmentation.

The comparison between MiniSeg and competitors, in terms of 
the number of parameters, the number of FLOPs, and speed, 
is summarized in \tabref{tab:compParameters}. 
We can clearly see that the numbers of parameters and FLOPs 
of MiniSeg are extremely small.
Meanwhile, the speed of MiniSeg is much faster than others.
The numerical evaluation results of MiniSeg and
other competitors are presented in \tabref{tab:comp}.
MiniSeg consistently achieves the best or close to the best
performance in terms of all metrics on all datasets.
For the metric of SPC, MiniSeg performs slightly worse than 
the best method on COVID-19-CT100 and COVID-19-P9.
On the COVID-19-P1110 dataset, MiniSeg does not achieve 
the best results in terms of SEN.
The fact that MiniSeg consistently outperforms other competitors
demonstrates its effectiveness and superiority
in COVID-19 infected area segmentation. 
Note that MiniSeg does not need to be pretrained 
on ImageNet \cite{russakovsky2015imagenet}
owing to its small model size.
Therefore, we can come to the conclusion that MiniSeg has 
a low computational load, a fast speed, and good accuracy,
making it convenient for practical deployment that is 
of high importance in the current severe situation of COVID-19.

\subsubsection{Qualitative Comparison.}
To explicitly show the superiority of MiniSeg, 
we provide a qualitative comparison between 
MiniSeg and eight \sArt methods in \figref{fig:cmp_sample}.
We select some representative images from the above datasets.
This visual comparison further indicates that 
MiniSeg outperforms baseline methods remarkably.

\subsubsection{Statistical Analysis.}
To further study the characteristics of MiniSeg, we perform statistical 
analysis on the largest COVID-19-P20 dataset.
\figref{fig:analysis_A} and \figref{fig:analysis_B} illustrate the relationship 
between the DSC score and the infected area, or the lesion count in a CT slice, 
respectively.
We find that MiniSeg achieves a DSC score larger than 0.7 for most CT slices
regardless of the infected area.
The medium DSC is above 0.8 regardless of the lesion counts.
This suggests that MiniSeg is robust to different cases for COVID-19 infected 
area segmentation.

\section{Conclusion}
In this paper, we focus on segmenting COVID-19 infected areas
from chest CT slices.
To address the lack of COVID-19 training data and meet the 
efficiency requirement for the deployment of computer-aided
COVID-19 screening systems, we propose an extremely minimum
network, \ie MiniSeg, for accurate and efficient COVID-19
infected area segmentation.
MiniSeg adopts a novel multi-scale learning module, 
\ie the \textbf{\underline{A}}ttentive 
\textbf{\underline{H}}ierarchical 
\textbf{\underline{S}}patial
\textbf{\underline{P}}yramid (\textbf{AHSP}) module,
to ensure its accuracy under the constraint of the extremely 
minimum network size.
To extensively compare MiniSeg with previous \sArt image 
segmentation methods and promote future research on COVID-19
infected area segmentation, we build a comprehensive benchmark 
that would be useful for future research.
The comparison between MiniSeg and \sArt image segmentation 
methods demonstrates that MiniSeg not only achieves the best 
performance but also has high efficiency, 
making MiniSeg suitable for practical deployment.

\bibliography{references}

\end{document}